\documentclass[letterpaper, 10 pt, conference]{ieeeconf}  

\usepackage{cite}

\usepackage{graphicx}
\DeclareGraphicsExtensions{.png}

\usepackage[cmex10]{amsmath}
\usepackage{amsfonts}
\usepackage{fixmath}

\usepackage{array}

\usepackage{caption}

\IEEEoverridecommandlockouts                              

\title{\LARGE \bf
Motion Imitation Based on Sparsely Sampled Correspondence
}

\author{Shuo Jin$^{1}$, Chengkai Dai$^{1}$, Yang Liu$^{2}$, Charlie C.L. Wang$^{3,*}$ 
\thanks{This work is supported by Hong Kong ITC Innovation and Technology Fund (ITS/065/14), and Chengkai Dai is partially supported by Hong Kong RGC General Research Fund (CUHK/14207414).}
\thanks{$^{1}$S. Jin and C. Dai are with the Department of Mechanical and Automation
Engineering, The Chinese University of Hong Kong, Hong Kong.}
\thanks{$^{2}$Y. Liu is with Microsoft Research Asia, Beijing, China.}
\thanks{$^{3}$C.C.L. Wang is with the Department of Design Engineering and TU Delft Robotics Institute, Delft University of Technology, The Netherlands.}%
\thanks{$^{*}$Corresponding Author. Email:{\tt\small c.c.wang@tudelft.nl}}%
}

\usepackage{color}

\begin{document}

\maketitle
\thispagestyle{empty}
\pagestyle{empty}

\begin{abstract}
Existing techniques for motion imitation often suffer a certain level of latency due to their computational overhead or a large set of correspondence samples to search. To achieve real-time imitation with small latency, we present a framework in this paper to reconstruct motion on humanoids based on sparsely sampled correspondence.
The imitation problem is formulated as finding the projection of a point from the configuration space of a human's poses into the configuration space of a humanoid. An optimal projection is defined as the one that minimizes a back-projected deviation among a group of candidates, which can be determined in a very efficient way. Benefited from this formulation, effective projections can be obtained by using sparse correspondence. Methods for generating these sparse correspondence samples have also been introduced.
Our method is evaluated by applying the human's motion captured by a RGB-D sensor to a humanoid in real-time. Continuous motion can be realized and used in the example application of tele-operation.
\end{abstract}

\section{Introduction}
Humanoid robots have been widely studied in the research of robotics. 
With the recent development of motion capture devices such as RGB-D camera
(e.g., Kinect) and wearable sensor system (e.g., Xsens MVN), efforts have
been made to generate human-like motions for humanoid robots with high degree-of-freedoms.
However, directly applying captured poses of human to humanoids 
is difficult because of the difference in human's and humanoid's kinematics. Therefore, a variety 
of kinematics based approaches for humanoid imitation have been investigated, which can be classified into two 
categories. Many of them perform an offline optimization step to compute the corresponding 
configurations that conform to the mechanical structures and kinematics of humanoids from input
human data\cite{suleiman2008human, chalodhorn2007learning, nakaoka2007learning, ude2004programming,
kim2009stable, safonova2003optimizing}. It is obvious that the significant computational overhead 
in those techniques prevents us from applying them to real-time imitation.
Methods in the other thread of research compute online imitation following captured human 
motion\cite{ott2008motion, dariush2009online, do2008imitation, 
yamane2010controlling, koenemann2012whole, koenemann2014real}.

In this paper, we consider about the problem of realizing real-time human-to-humanoid motion imitation.
Unfortunately, it is not an easy task due to:
\begin{itemize}
\item full sampling of human-to-humanoid correspondence often leads to large data size;

\item high non-linearity of underlying mechanical rules results in significant computational cost;

\item how to find the configuration of a humanoid according to the input poses of human in real-time is not intuitive.
\end{itemize}
Artificial neural networks have been adopted to ease the difficulties, with which a lot of
efforts have been made in simulation and for robots with small degree-of-freedoms\cite{morris1997finding,
aleotti2004position, neto2009accelerometer, neto2010high, 
stanton2012teleoperation, van1993control, jung1996neural, larsen2004case, wang2005improving}. 
A recent work\cite{stanton2012teleoperation} by
Stanton et al directly introduced neural networks with particle swarm optimization to find the
mapping between human movements and joint angle positions of humanoid. However, there is no measurement presented in their work to evaluate the quality of humanoid poses generated by the trained neural system. On the other aspect, our method is also different from this work in terms of the training data set. We use the sparse correspondence 
instead of the densely recorded raw data, which can help eliminate the redundancy in data set and 
improve the training speed. Moreover, only requiring a sparse set of correspondence samples leads to a lower barrier of system implementation. 

\begin{figure}\centering
\includegraphics[width=\linewidth]{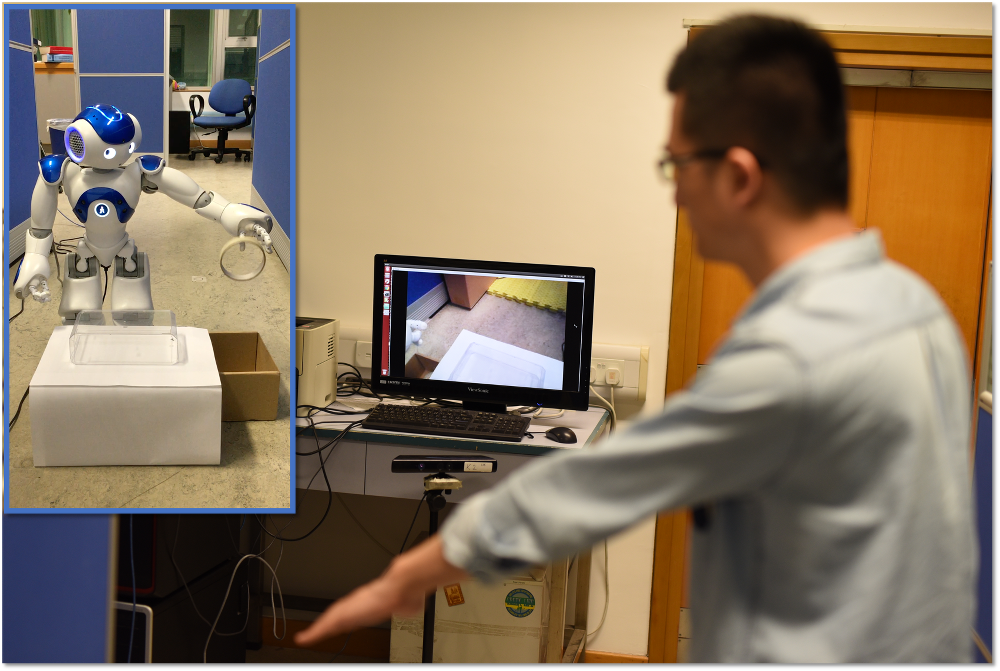}
\caption{An example of humanoid imitation realized by our framework.}
\label{figTeleoperation}
\end{figure}

\begin{figure*} \centering
\captionsetup{justification=centering}
\includegraphics[width=\linewidth]{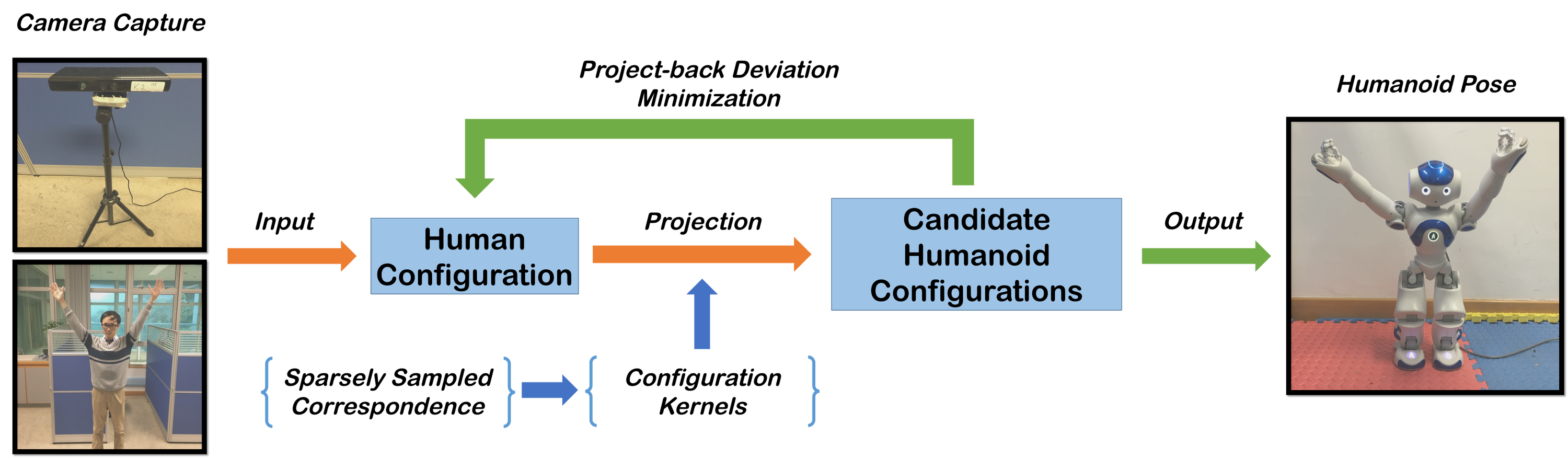}
\caption{An illustration of our framework for motion imitation using configuration projection.}
\label{figWorkflow}
\end{figure*}

We propose a framework that allows efficient projection of a pose from
human's space to the configuration space of humanoid based on sparsely sampled correspondence 
extracted from recorded raw data, which can be used to realize motion imitation in real time (see Fig. \ref{figWorkflow}). 
Experimental results show that our framework can be successfully used in the motion imitation of humanoid (see Fig.\ref{figTeleoperation} for an example of tele-operation using a NAO humanoid).

\section{Framework of Configuration Projection} \label{secFramework}
\subsection{Problem Definition} \label{subsecProblem}
A human pose can be uniquely represented as a point~(abbreviated as $C$-point) $\mathbf{h} \in \mathbb{R}^m$ 
in the configuration space (abbreviated as $C$-space -- $\mathcal{H}$) of human's motion and its corresponding pose of humanoid can be denoted as a point $\mathbf{r} \in \mathbb{R}^n$ in the $C$-space of humanoid -- $\mathcal{R}$. We assume one-to-one correspondence between the poses of human body and humanoid,
i.e. the mapping between human and humanoid's $C$-spaces is bijective. A pair of
human's and humanoid's configurations is denoted as $(\mathbf{h}, \mathbf{r}) \in \mathbb{R}^{m+n}$. 
Given stored correspondence pairs $\{(\mathbf{h}, \mathbf{r})\}$ as the known knowledge and a new input pose $\mathbf{h}^*\in \mathbb{R}^m$, the configuration projection $\Omega(\cdot)$ can be
defined as finding a corresponding $\mathbf{r}^* \in \mathbb{R}^n$ that satisfies two basic properties:
\begin{itemize}
\item \textbf{Identity} -- for any sample pair $(\mathbf{h}_i, \mathbf{r}_i)$ in the data-set, it should have
\begin{center}
$\Omega(\mathbf{h}_i) = \mathbf{r}_i$.
\end{center}

\item \textbf{Similarity} -- for an input $C$-point of human $\mathbf{h}^*$, if 
\begin{displaymath}
\max \{ \min_i \|\mathbf{h}_i - \mathbf{h}^*\| \} < \delta
\end{displaymath}
then it should have
\begin{displaymath}
\|\Omega(\mathbf{h}^*) - \tilde{\mathbf{r}}(\mathbf{h}^*)\| < \epsilon,
\end{displaymath}
where $\delta$ and $\epsilon$ are two constant values, and $\tilde{\mathbf{r}}(\mathbf{h}^*)$ is a $C$-point of humanoid that 
can be obtained by more accurate but computational intensive methods (e.g., inverse kinematics) as the ground truth. 
\end{itemize}
All sample pairs should be repeated with the projection $\Omega(\cdot)$ according to the property of \textit{identity}.
The demand on \textit{similarity} indicates that
if a new input is close to the known samples, 
its projected result should not deviate too much from its corresponding ground truth.

The main difficulty of finding the projection $\mathbf{r}^*$ lies in the lack of explicit functions 
to determine the mapping between two $C$-spaces with different dimensions (i.e., degree-of-freedoms). 
Given sparsely aligned pairs of poses as samples, 
we try to solve this problem by proposing a strategy of kernel-based projection to
find a good approximation for $\mathbf{r}^*$.

\subsection{Data Pre-processing}\label{subsecPreprocessing}
The knowledge of correspondence $\{(\mathbf{h}, \mathbf{r})\}$ can be established through experiments. Although aligning a pose of human body with a corresponding pose of humanoid can be taken manually, it is a task almost impossible if thousands of such correspondence samples need to be specified. Therefore, in our experiments, we first capture continuous motions of human bodies by using a motion capture system. The data-set obtained in this way often results in
large size and redundancy. To resolve this problem, we perform a pre-processing step to extract marker poses from the raw data-set recorded from human's motion. Specifically, mean shift clustering~\cite{comaniciu2002meanshift} is employed to
generate the marker set denoted as $\mathcal{H}$. For each sample $\hat{\mathbf{h}}\in \mathcal{H}$, 
its corresponding pose $\hat{\mathbf{r}}$ in the configuration space of humanoid can be either specified manually (when the number of samples in $\mathcal{H}$ is small) or generated automatically by a sophisticated method (e.g., the inverse kinematics methods). The pairs of correspondence, $\{(\hat{\mathbf{h}}_i,\hat{\mathbf{r}}_i) \}_{i=1,\cdots,N}$, extracted in this way is treated as landmarks to be used in our framework.

\subsection{ELM Based Kernels}
As the configuration pairs of marker data-set are discrete in space, we define a kernel $\kappa(\cdot)$
on each marker configuration $\hat{\mathbf{h}}_i$ and $\hat{\mathbf{r}}_i$ 
as a local spatial descriptor using the technique of \textit{Extreme Learning Machine} (ELM) \cite{huang2011extreme}. 
ELM method has been widely used in regression and classification
problems as a \textit{single hidden layer feed-forward network} (SLFN) with its advantageous properties of fast training
speed, tuning-free neurons and easiness in implementation (ref.~\cite{huang2012extreme}).
Basically, the training formula of ELM can be expressed as $\mathbf{H} \mathbf{b} = \mathbf{T}$,
where $\mathbf{H}$ is the hidden layer output matrix of SLFN, $\mathbf{b}$ is the output weight
vector to be computed, and $\mathbf{T}$ is the target feature vector. 

Given a new input $\mathbf{x}$,
the prediction function of ELM is $\mathbf{f}(\mathbf{x}) = \mathbf{Q}(\mathbf{x})\mathbf{b}$,
where the $\mathbf{Q}(\mathbf{x})$ is the hidden layer feature mapping of $\mathbf{x}$. It has been pointed
out in~\cite{huang2011extreme} that the training errors will be eliminated if the number of hidden nodes is
not less than the number of training samples, indicating the trained ELM can be used as a fitting function that
interpolates all training samples
\begin{displaymath}
\mathbf{Q}(\hat{\mathbf{h}}_i)\mathbf{b} = \hat{\mathbf{r}}_i, ~~ (i = 1, \cdots, N).
\end{displaymath}
In this case, the output weight vector is computed as
\begin{displaymath}
\mathbf{b} = \mathbf{H}^{T}(\mathbf{H}\mathbf{H}^{T})^{-1}\mathbf{T},
\end{displaymath}
where $\mathbf{H}^{T}(\mathbf{H}\mathbf{H}^{T})^{-1}$ is the Moore-Penrose generalized inverse of
$\mathbf{H}$. Regularized ELM is proposed in\cite{deng2009regularized} to improve its numerical
stability, leading to the following training formula with $\lambda$ (a very small value in practice) 
as the regularization factor
\begin{displaymath}
\mathbf{b} = \mathbf{H}^{T}(\lambda + \mathbf{H}\mathbf{H}^{T})^{-1}\mathbf{T}.
\end{displaymath}

With the help of ELM, a kernel $\kappa^h_{i}(\cdot) \in \mathbb{R}^n$ for a human's landmark point $\hat{\mathbf{h}}_i$ can be built with its nearest neighbors. 
Specifically, we find $k$ spatial nearest neighbors of $\hat{\mathbf{h}}_i$ in the set of human's landmarks as 
$\{ \hat{\mathbf{h}}_j \}_{j \in \mathcal{N}(\hat{\mathbf{h}}_i)}$, where $\mathcal{N}(\cdot)$ denotes the set of nearest neighbors. Then, the ELM kernel of $\kappa^h_{i}(\cdot)$ is trained using $\{ ( \hat{\mathbf{h}}_j, \hat{\mathbf{r}}_j ) \}_{j \in \mathcal{N}(\hat{\mathbf{h}}_i)}$, which is regarded as an approximate local descriptor of the nearby mapping of $\hat{\mathbf{h}}_i$: $\mathcal{H} \mapsto \mathcal{R}$. When inputting a new human pose $\mathbf{h}^* \in \mathbb{R}^m$, a local estimation of mapping with reference to this kernel can be represented as
\begin{displaymath}
\kappa^h_{i}(\mathbf{h}^*)=\mathbf{Q}(\mathbf{h}^*) \mathbf{b}.
\end{displaymath}
This function is called a \textit{forward} kernel.
Similarly, for each $C$-point ${r^m}_i$ of a humanoid, an ELM based kernel $\kappa^r_{i}(\cdot) \in \mathbb{R}^m$ can be constructed in the same way for the inverse mapping: $\mathcal{R} \mapsto \mathcal{H}$. $\kappa^r_{i}(\cdot)$ is called a \textit{backward} kernel. These two types of kernel functions will be used in our framework for realizing the projection.

\subsection{Projection}\label{subsecFramework}
For an input pose $\mathbf{h}^* \in \mathbb{R}^m$, the point determined by the ELM kernel function, $\kappa^h_{i}(\mathbf{h}^*)$, is not guaranteed to satisfy the requirement of bijective mapping (i.e., $\kappa^r_{i}(\kappa^h_{i}(\mathbf{h}^*)) \neq \mathbf{h}^*$). To improve the bijection of mapping, the projection of a human's $C$-point is formulated as determining an optimal point from all candidates generated from different forward kernels. 
\begin{figure}[t] \centering
\includegraphics[width=\linewidth]{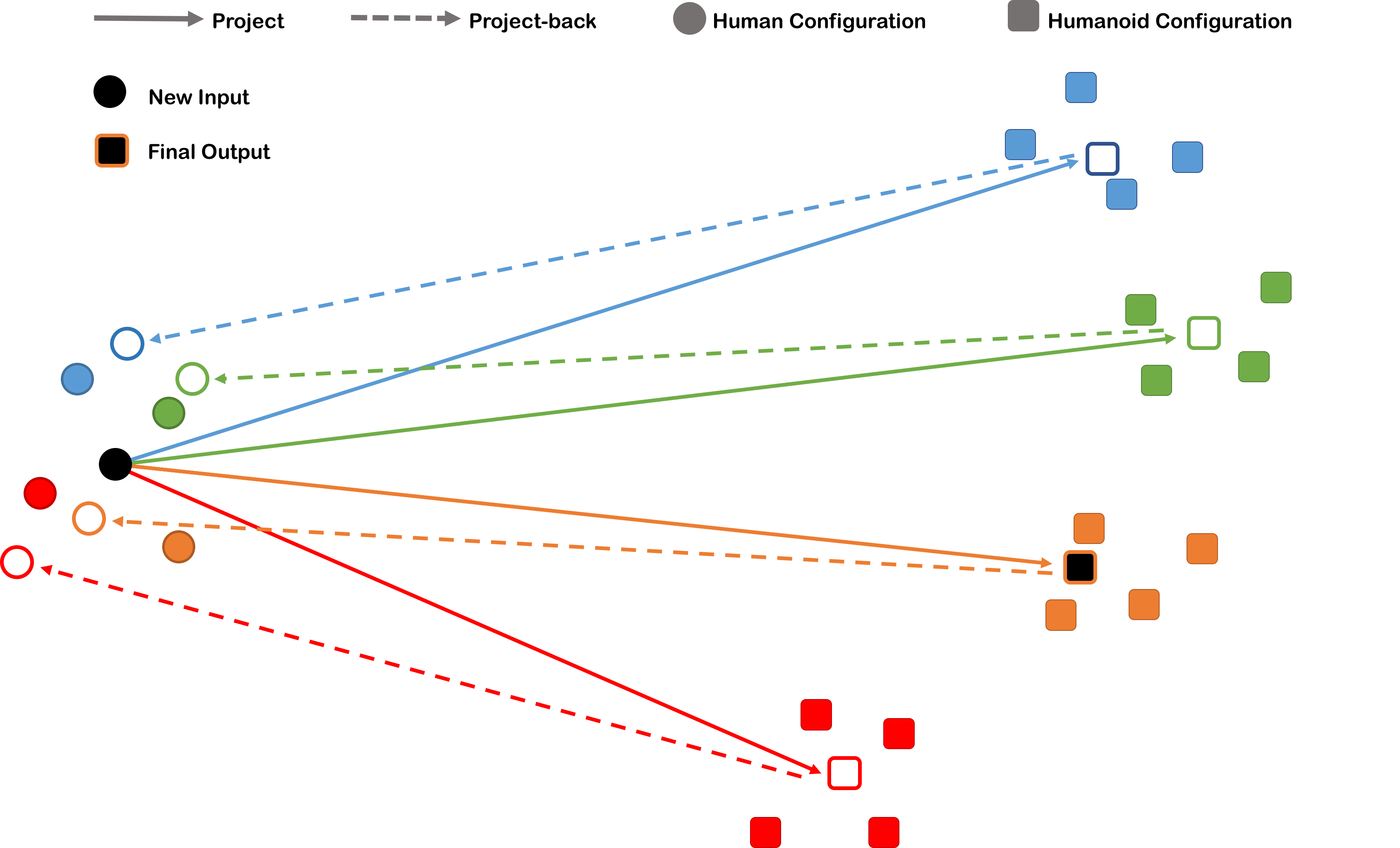}
\caption{An illustration of finding an optimal point that minimizes a back-projected deviation (with $L=M=4$). 
}
\label{figFramework}
\end{figure}

First of all, $L$ nearest neighbors of $\mathbf{h}^*$ are retrieved in $\mathcal{H}$ as $\{\hat{\mathbf{h}}_j\}$ ($j=1,\cdots,L$). From the forward kernel associated with each of these $L$ points in $\mathcal{H}$, a candidate point in $\mathcal{R}$ can be determined by $\mathbf{r}^c_j=\kappa^h_j(\mathbf{h}^*)$. For each $\mathbf{r}^c_j$, we search for its $M$ nearest neighbors in $\mathcal{R}$ as $\mathcal{N}(\mathbf{r}^c_j)=\{\hat{\mathbf{r}}_{j,k}\}$ ($k=1,\cdots,M$). In other words, there are $M$ backward kernels associated with $\mathbf{r}^c_j$, which are $\{ \kappa^r_{j,k}\}$. In each cluster of backward kernels, we determine a set of weights $w_{j,k}$ that leads to a point formed as the convex combination of $\{\hat{\mathbf{r}}_{j,k}\}$ 
\begin{displaymath}
\tilde{\mathbf{r}}^c_j = \sum_{k} w_{j,k} \mathbf{r}^c_{j,k}.
\end{displaymath}
An optimal point $\tilde{\mathbf{r}}^c_j$ minimizes the deviation of back-projection with regard to the cluster of kernels $\{\kappa^r_{j,k}(\cdot)\}_k$ is defined as 
\begin{equation} \label{eqSubFunction}
\begin{aligned}
			&\min_{w_{j,k}} \{ \| \kappa^r_{j,k} (\sum_{k} w_{j,k} \mathbf{r}^c_{j,k}) - \mathbf{h}^* \| \}_k, \\ 
s.t. \quad  & \sum_{k=1}^M w_{j,k} = 1, ~~ w_{j,k} \geq 0.
\end{aligned}
\end{equation}
The final projected point $\mathbf{r}^*$ is then defined as 
\begin{equation} \label{eqCoreFunction}
\mathbf{r}^* = \sum_{k} w_{l,k} \mathbf{r}^c_{l,k}
\end{equation}
according to the cluster of $\mathcal{N}(\mathbf{r}^c_l)$ that gives the minimal back-projected deviation, which is a solution of
\begin{equation}\begin{aligned}\label{eqCoreFunction2}
			& \min_j \left\{ \min_{w_{j,k}} \{ \| \kappa^r_{j,k} (\sum_{k} w_{j,k}\mathbf{r}^c_{j,k}) - \mathbf{h}^* \| \}_k \right\}, \\ 
s.t. \quad  &  \sum_{k=1}^M w_{j,k} = 1, ~~ w_{j,k} \geq 0.
\end{aligned}\end{equation} 
The computation for solving above optimization problem can be slow in many cases. Therefore, we propose a sub-optimal objective function as a relaxation of Eq.(\ref{eqCoreFunction2}) to be used in real-time applications (e.g., the tele-operation shown in Fig.\ref{figTeleoperation}). The problem is relaxed to
\begin{equation}
\min_j \left\{ \min_{k} \{ \| \kappa^r_{j,k} (\mathbf{r}^c_{j}) - \mathbf{h}^* \| \}_k \right\},
\end{equation} 
the solution of which can be acquired very efficiently by checking each candidate $\mathbf{r}^c_j$ with 
regard to all its $M$ reference backward kernels. Figure \ref{figFramework} gives an illustration for the evaluation of back-projected deviation.

\vspace{5pt} \noindent \textbf{Motion Smoothing:} A dynamic motion is processed as a sequence of continuous poses
in our system, where the projected poses in the configuration space of humanoid are generated separately. To avoid the generation of jerky motion, we use a method modified from the double exponential smoothing~\cite{laviola2003double} to post-process the projected poses. Given a projected pose $\mathbf{r}_t$ at time frame $t$, the update rules of a smoothed pose $\mathbf{s}_t$ are defined as
\begin{equation}
\begin{aligned}
& \mathbf{s}_t = \alpha y_t + (1-\alpha)(\mathbf{s}_{t-1} + \mathbf{b}_{t-1}), ~~ 0 \leq \alpha \leq 1 \\
& \mathbf{b}_t = \gamma (\mathbf{s}_t - \mathbf{s}_{t-1}) + (1 - \gamma) \mathbf{b}_{t-1}, ~~ 0 \leq \gamma \leq 1 \\
& \mathbf{s}_t = \mathbf{s}_{t-1}, ~\text{if} ~\|\mathbf{s}_t - \mathbf{s}_{t-1}\| < \eta
\end{aligned}
\end{equation}
$\alpha$, $\gamma$ and $\eta$ are parameters to control the effectiveness of smoothing, where $\alpha=0.75$, $\gamma=0.3$ and $\eta=0.15$ are used to give satisfactory results in our practice.

\vspace{5pt} \noindent \textbf{Remark:} It must be clarified the \textit{Identity} property introduced in Section
\ref{subsecProblem} is relaxed to $\Omega(h_i) \approx r_i$ in practice due to the following reasons:
\begin{itemize}
\item Regularized ELM method is employed to construct the kernels, which changes the corresponding
energy function where a regularization term is added to improve its numerical stability.

\item Double exponential smoothing is applied for smoothing a motion, which introduces minor adjustments
on the output values.
\end{itemize}

\begin{figure} \centering
\includegraphics[width=0.9\linewidth]{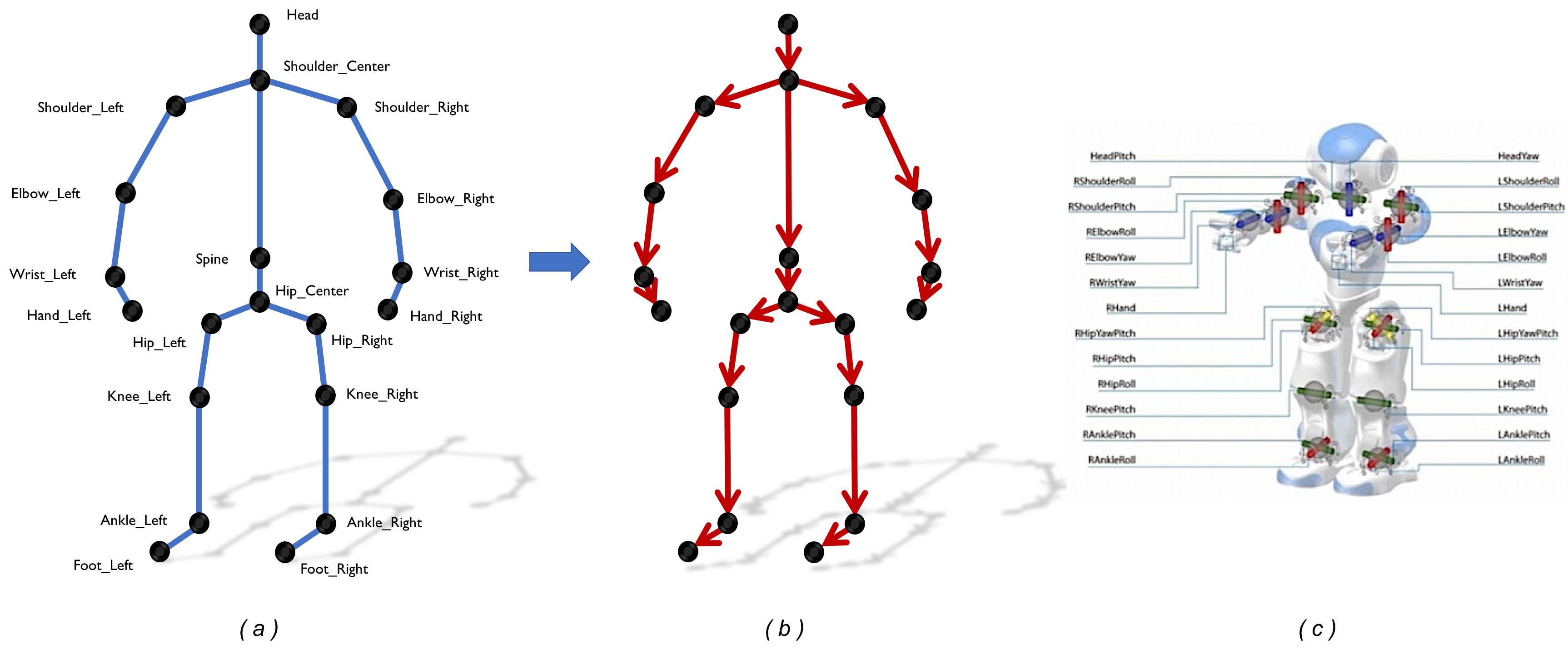}
\caption{Feature vectors of human and humanoid: (a) the human skeleton from a Kinect sensor,
(b) the corresponding pose descriptor of a human body consists of $19$ unit vectors, and
(c) the pose descriptor for a NAO humanoid formed by all DOFs on its joints (source: http//:www.ez-robot.com).}
\label{figSkeleton}
\end{figure}
\section{Real-time Projection on NAO} \label{secReconstruction}
Our framework is testified on real-time motion imitation of a NAO humanoid robot with a Kinect RGB-D camera as the device to capture the motion of human.

\subsection{Human-to-humanoid Motion Imitation}
The human skeleton provided by a Kinect sensor is a set of line segments based on predefined key joints
as shown in Fig.\ref{figSkeleton}(a). 
We define an abstraction consisting of $19$ unit vectors for a pose
as illustrated in Fig.\ref{figSkeleton}(b), which is independent different body dimensions. 
It should be pointed out that it is unnecessary to always use the full set of unit vectors unless full
body motion must be sensed. The NAO humanoid robot has 26 degree-of-freedoms, including the roll, 
pitch, and yaw of all its joints (see Fig.\ref{figSkeleton}(c)). 
Posing a NAO humanoid can be executed by specifying the values of all its degree-of-freedoms.

\begin{figure}[t]\centering
\includegraphics[width=\linewidth]{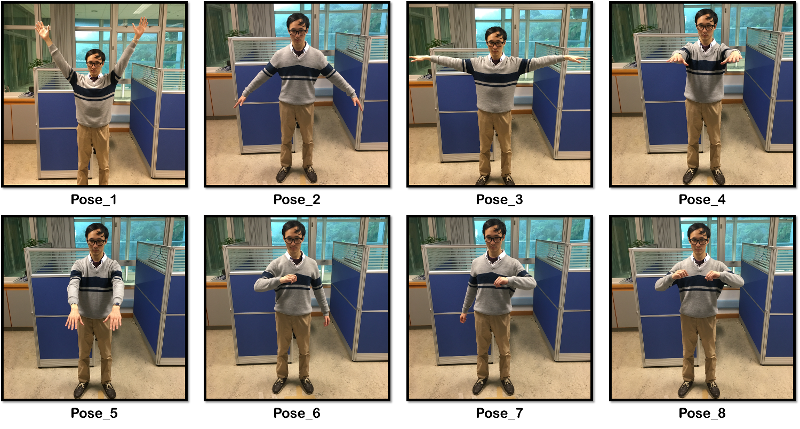}
\caption{Basic poses serve as benchmarks for similarity evaluation.}\label{figBasicPoses}
\end{figure}

To collect the data-set of human's motion, a user is asked to do arbitrary motion in front of a Kinect camera.
Meanwhile, we have implemented a straightforward \textit{inverse kinematics} (IK) based scheme for upper-body motion. 
The roll, pitch, and yaw of every joint can be computed directly by the unit vectors of a human's skeleton model. 
After using mean shift to extract the landmarks of motion from the raw set, their corresponding landmark poses in the $C$-space of humanoid can be generated by this IK. Besides, we also define \textit{eight} basic poses (see Fig.\ref{figBasicPoses}) which play a critical role when evaluating the similarity between the projected poses of humanoid and the poses of human.

Using the landmark poses defined in this way, human-to-humanoid motion imitation has been implemented by a single-core C++ program. All the tests below are taken on a personal computer with Intel Core i7-3770 3.4 GHz and 8 GB RAM memory. 

\subsection{Experimental Results}
We evaluate our method mainly from three perspectives, including the computational
efficiency of projection, the quality of reconstructed motion, and the influence by the size
of landmark set. 

\vspace{5pt}\noindent \textbf{Efficiency of Projection:} From Section \ref{subsecFramework}, we know that the complexity for computing projection depends on the size of neighbors (i.e., $L$ and $M$). 
The cost of computation increases with larger $L$ and $M$ as more candidates and more reference
kernels will be involved. In all our experiments, we use $L = M = 10$ and the average time for 
making a configuration projection is $0.00273ms$. When increasing to $L = M = 50$, the average time cost
is still only $0.0201ms$. In summary, the overhead of our method for motion imitation is very light -- i.e., it fits well for different real-time applications.

\begin{figure*} \centering
\includegraphics[width=\linewidth]{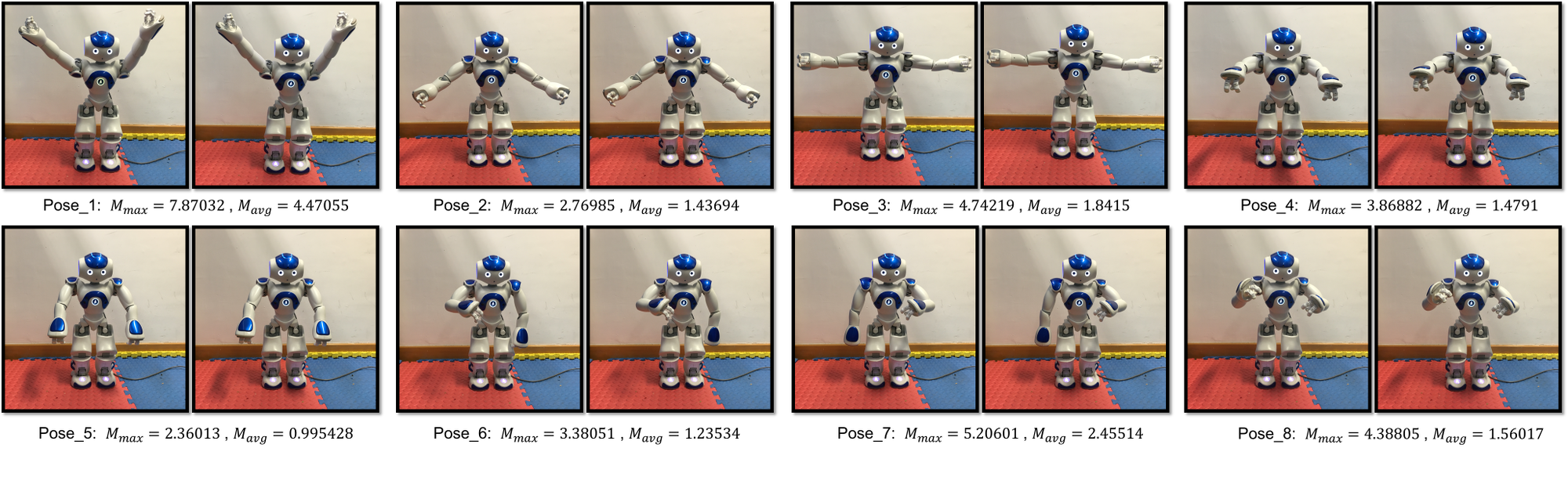}
\caption{Eight basic poses are reconstructed by our method (left of each pair) and compared with the ground truth (right of each pair). The similarity metrics, $M_{max}$ and $M_{avg}$, of each pair are also reported. The evaluation is taken on a projection defined by using $1,644$ landmark pairs.}\label{figPoseReconstruct}
\end{figure*}
\begin{figure*} \centering
\includegraphics[width=\linewidth]{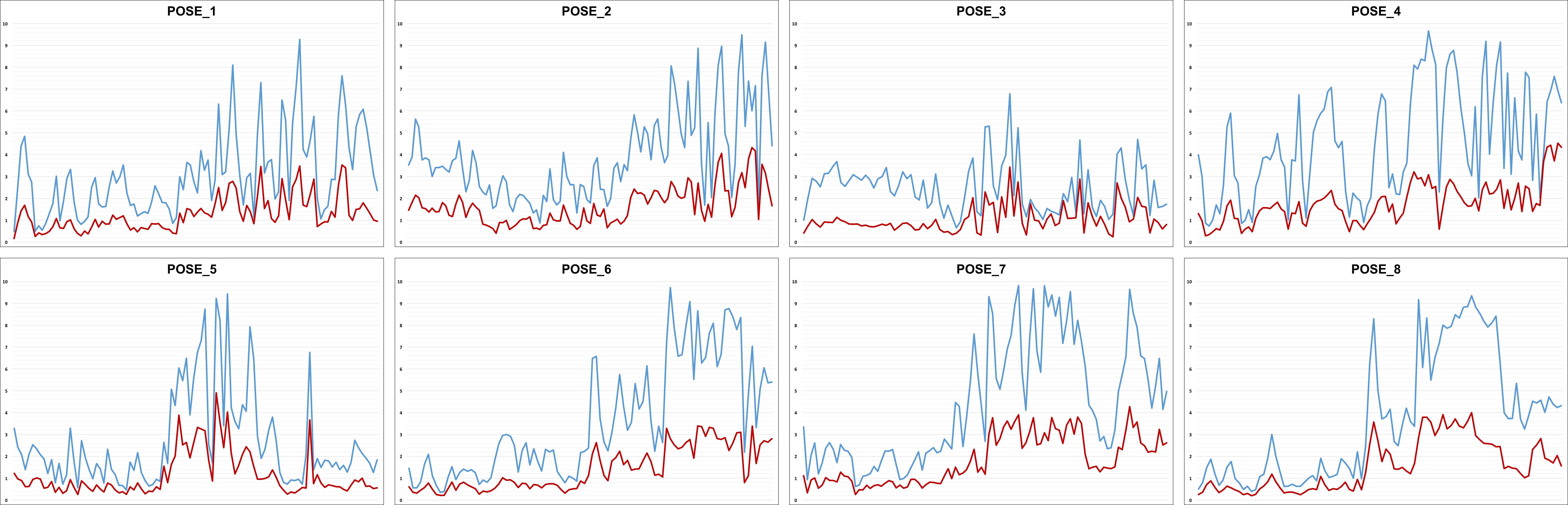}
\caption{Statistics in eight motions for the change of two metrics: $M_{max}$ (blue) and $M_{avg}$ (red). The evaluation is also taken on a projection with $1,644$ landmark pairs.}
\label{figStatistics}
\end{figure*}

\vspace{5pt}\noindent \textbf{Quality of Reconstruction:} Two metrics are used in our experiments to estimate the quality of a projected configuration $\mathbf{r}^* \in \mathbb{R}^{n}$ referring to its corresponding ground truth value $\mathbf{r}_{gt}$ -- the maximum absolute deviation in degree as
\begin{displaymath}
M_{max} = \frac{180^{\circ}}{\pi} \|r^* - r_{gt}\|_{\infty},
\end{displaymath}
and the average absolute deviation in degree as
\begin{displaymath}
M_{avg} = \frac{1}{n} \left( \frac{180^{\circ}}{\pi}\|r^* - r_{gt}\|_1 \right).
\end{displaymath}
The evaluation is taken with a set holding $1,644$ configuration pairs as landmarks. All those eight poses shown in Fig.\ref{figBasicPoses} are tested, and the results are shown in Fig.\ref{figPoseReconstruct}. The results of  comparison (in terms of $M_{max}$ and $M_{avg}$) indicates that the poses generated by our method share good similarity with the ground truths. Besides of static poses, we also evaluate the quality of reconstructed motion in the $C$-space of humanoid as a sequence of poses. We define eight basic motion sequences, each of which starts from the rest pose and ends at one of the basic poses. The complete human motions are recorded for the reconstruction using our projection in the $C$-space of humanoid. The projected poses are compared with the poses generated by IK, serving as the ground truths. The values of $M_{max}$ and $M_{avg}$ in these eight motions are shown in Fig.\ref{figStatistics}. It is easy to find that the errors are bounded to less than $10^{\circ}$ in all motions. 
\begin{figure}[t] \centering
\includegraphics[width=\linewidth]{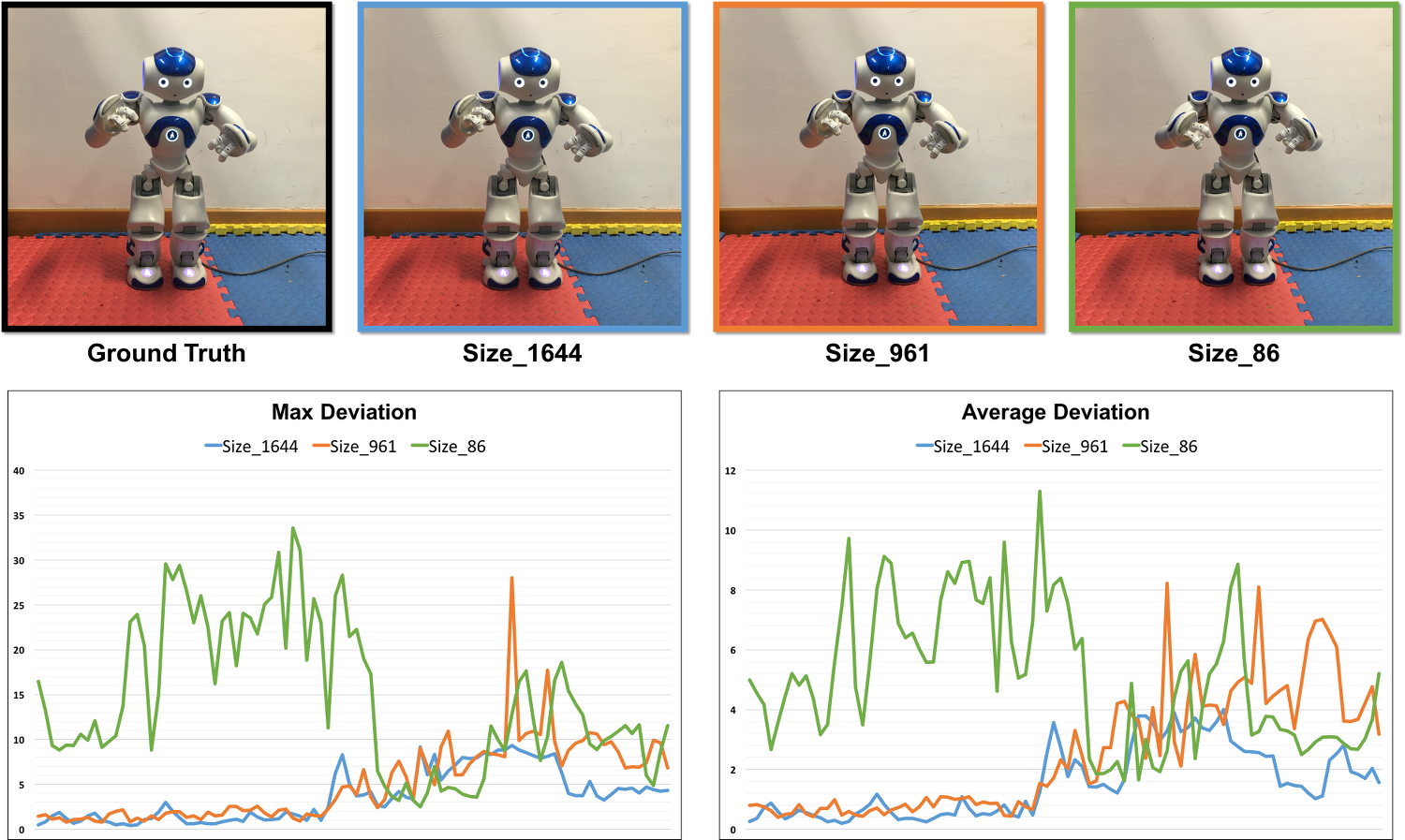}
\caption{To reconstruct motion using landmark sets having different number of corresponding samples, statistics indicate that more landmark pairs lead to better results.}\label{figSize}
\end{figure}

\vspace{5pt}\noindent \textbf{Size of Landmarks:} As presented in Section \ref{subsecPreprocessing}, the correspondence samples used to formulate projection in our framework is extracted from the captured motions. In our implementation, it is generated by a user moving in front of a Kinect sensor for 5 minutes. Then, three sets with different number of landmarks ($1,644$, $961$, and $86$ respectively) are extracted. The corresponding pairs of poses are then constructed with the help of IK. The 8-th pose in Fig.\ref{figBasicPoses} -- POSE\_8 and the motion from the rest pose to POSE\_8 are constructed from the projections defined on the sets with different number of landmarks. From the statistics and comparisons shown in Fig.\ref{figSize}, it is easy to conclude that our projection based formulation converges when the number of landmarks increases. In other words, more landmarks result in a more accurate projection. However, it should also be noted that the projection from the smaller set may still be useful in some applications with low requirement on quality but having more restrictions on speed and memory usage.

\subsection{Application of tele-operation}
We have tested the motion imitation realized by our method in an application of tele-operation using a NAO humanoid.
As illustrated in Fig.\ref{figTeleoperation} and the supplementary video of this paper, a user can remotely control the motion of a NAO robot to grasp an object and put it into a box. The scene that can be seen from the camera of NAO is displayed on a screen placed in front of the user as the visual feedback. The imitation realized by our system has good accuracy. As a result, the tele-operation can be performed very smoothly.

\section{Conclusion \& Future Work} \label{secConclusion}
In this paper, we have proposed a framework to realize motion imitation. Different from conventional methods, our method is based on a novel formulation of projection between two configuration spaces with different dimensions. Given a new input pose of human, its projection in the space of humanoid is defined as finding the optimal $C$-point that minimizes a back-projection deviation referring to the built kernels. We have validated our idea by reconstructing humanoid motion on a NAO robot. The experimental results are encouraging and motions of good quality can be reconstructed efficiently.

There are several potential improvements can be made to our method. 
First, an intuitive improvement is to extend the current setup to full-body motion reconstruction by 
incorporating the constraint of whole-body balance.
Second, the ELM based kernels currently used in our framework do not have a explicit bound for prediction
with a new input. Finding kernel functions that can provide a bound on prediction could be another future work.
Lastly, we are interested in exploring more applications beyond tele-operation.


\bibliographystyle{IEEEtran}
\bibliography{IROSbibliography}

\begin{thebibliography}{10}
\providecommand{\url}[1]{#1}
\csname url@samestyle\endcsname
\providecommand{\newblock}{\relax}
\providecommand{\bibinfo}[2]{#2}
\providecommand{\BIBentrySTDinterwordspacing}{\spaceskip=0pt\relax}
\providecommand{\BIBentryALTinterwordstretchfactor}{4}
\providecommand{\BIBentryALTinterwordspacing}{\spaceskip=\fontdimen2\font plus
\BIBentryALTinterwordstretchfactor\fontdimen3\font minus
  \fontdimen4\font\relax}
\providecommand{\BIBforeignlanguage}[2]{{%
\expandafter\ifx\csname l@#1\endcsname\relax
\typeout{** WARNING: IEEEtran.bst: No hyphenation pattern has been}%
\typeout{** loaded for the language `#1'. Using the pattern for}%
\typeout{** the default language instead.}%
\else
\language=\csname l@#1\endcsname
\fi
#2}}
\providecommand{\BIBdecl}{\relax}
\BIBdecl

\bibitem{suleiman2008human}
W.~Suleiman, E.~Yoshida, F.~Kanehiro, J.-P. Laumond, and A.~Monin, ``On human
  motion imitation by humanoid robot,'' in \emph{Robotics and Automation, 2008.
  ICRA 2008. IEEE International Conference on}.\hskip 1em plus 0.5em minus
  0.4em\relax IEEE, 2008, pp. 2697--2704.

\bibitem{chalodhorn2007learning}
R.~Chalodhorn, D.~B. Grimes, K.~Grochow, and R.~P. Rao, ``Learning to walk
  through imitation.'' in \emph{IJCAI}, vol.~7, 2007, pp. 2084--2090.

\bibitem{nakaoka2007learning}
S.~Nakaoka, A.~Nakazawa, F.~Kanehiro, K.~Kaneko, M.~Morisawa, H.~Hirukawa, and
  K.~Ikeuchi, ``Learning from observation paradigm: Leg task models for
  enabling a biped humanoid robot to imitate human dances,'' \emph{The
  International Journal of Robotics Research}, vol.~26, no.~8, pp. 829--844,
  2007.

\bibitem{ude2004programming}
A.~Ude, C.~G. Atkeson, and M.~Riley, ``Programming full-body movements for
  humanoid robots by observation,'' \emph{Robotics and autonomous systems},
  vol.~47, no.~2, pp. 93--108, 2004.

\bibitem{kim2009stable}
S.~Kim, C.~Kim, B.~You, and S.~Oh, ``Stable whole-body motion generation for
  humanoid robots to imitate human motions,'' in \emph{Intelligent Robots and
  Systems, 2009. IROS 2009. IEEE/RSJ International Conference on}.\hskip 1em
  plus 0.5em minus 0.4em\relax IEEE, 2009, pp. 2518--2524.

\bibitem{safonova2003optimizing}
A.~Safonova, N.~Pollard, and J.~K. Hodgins, ``Optimizing human motion for the
  control of a humanoid robot,'' \emph{Proc. Applied Mathematics and
  Applications of Mathematics}, vol.~78, 2003.

\bibitem{ott2008motion}
C.~Ott, D.~Lee, and Y.~Nakamura, ``Motion capture based human motion
  recognition and imitation by direct marker control,'' in \emph{Humanoid
  Robots, 2008. Humanoids 2008. 8th IEEE-RAS International Conference
  on}.\hskip 1em plus 0.5em minus 0.4em\relax IEEE, 2008, pp. 399--405.

\bibitem{dariush2009online}
B.~Dariush, M.~Gienger, A.~Arumbakkam, Y.~Zhu, B.~Jian, K.~Fujimura, and
  C.~Goerick, ``Online transfer of human motion to humanoids,''
  \emph{International Journal of Humanoid Robotics}, vol.~6, no.~02, pp.
  265--289, 2009.

\bibitem{do2008imitation}
M.~Do, P.~Azad, T.~Asfour, and R.~Dillmann, ``Imitation of human motion on a
  humanoid robot using non-linear optimization,'' in \emph{Humanoid Robots,
  2008. Humanoids 2008. 8th IEEE-RAS International Conference on}.\hskip 1em
  plus 0.5em minus 0.4em\relax IEEE, 2008, pp. 545--552.

\bibitem{yamane2010controlling}
K.~Yamane, S.~O. Anderson, and J.~K. Hodgins, ``Controlling humanoid robots
  with human motion data: Experimental validation,'' in \emph{Humanoid Robots
  (Humanoids), 2010 10th IEEE-RAS International Conference on}.\hskip 1em plus
  0.5em minus 0.4em\relax IEEE, 2010, pp. 504--510.

\bibitem{koenemann2012whole}
J.~Koenemann and M.~Bennewitz, ``Whole-body imitation of human motions with a
  nao humanoid,'' in \emph{Human-Robot Interaction (HRI), 2012 7th ACM/IEEE
  International Conference on}.\hskip 1em plus 0.5em minus 0.4em\relax IEEE,
  2012, pp. 425--425.

\bibitem{koenemann2014real}
J.~Koenemann, F.~Burget, and M.~Bennewitz, ``Real-time imitation of human
  whole-body motions by humanoids,'' in \emph{Robotics and Automation (ICRA),
  2014 IEEE International Conference on}.\hskip 1em plus 0.5em minus
  0.4em\relax IEEE, 2014, pp. 2806--2812.

\bibitem{morris1997finding}
A.~S. Morris and A.~Mansor, ``Finding the inverse kinematics of manipulator arm
  using artificial neural network with lookup table,'' \emph{Robotica},
  vol.~15, no.~06, pp. 617--625, 1997.

\bibitem{aleotti2004position}
J.~Aleotti, A.~Skoglund, and T.~Duckett, ``Position teaching of a robot arm by
  demonstration with a wearable input device,'' in \emph{International
  Conference on Intelligent Manipulation and Grasping (IMG04)}, 2004, pp. 1--2.

\bibitem{neto2009accelerometer}
P.~Neto, J.~N. Pires, and A.~P. Moreira, ``Accelerometer-based control of an
  industrial robotic arm,'' in \emph{Robot and Human Interactive Communication,
  2009. RO-MAN 2009. The 18th IEEE International Symposium on}.\hskip 1em plus
  0.5em minus 0.4em\relax IEEE, 2009, pp. 1192--1197.

\bibitem{neto2010high}
P.~Neto, J.~Norberto~Pires, and A.~Paulo~Moreira, ``High-level programming and
  control for industrial robotics: using a hand-held accelerometer-based input
  device for gesture and posture recognition,'' \emph{Industrial Robot: An
  International Journal}, vol.~37, no.~2, pp. 137--147, 2010.

\bibitem{stanton2012teleoperation}
C.~Stanton, A.~Bogdanovych, and E.~Ratanasena, ``Teleoperation of a humanoid
  robot using full-body motion capture, example movements, and machine
  learning,'' in \emph{Proceedings of Australasian Conference on Robotics and
  Automation}.\hskip 1em plus 0.5em minus 0.4em\relax Victoria University of
  Wellington New Zealand, 2012, pp. 3--5.

\bibitem{van1993control}
P.~Van~der Smagt and K.~Schulten, ``Control of pneumatic robot arm dynamics by
  a neural network,'' in \emph{Proc. of the 1993 World Congress on Neural
  Networks}, vol.~3.\hskip 1em plus 0.5em minus 0.4em\relax Citeseer, 1993, pp.
  180--183.

\bibitem{jung1996neural}
S.~Jung and T.~Hsia, ``Neural network reference compensation technique for
  position control of robot manipulators,'' in \emph{Neural Networks, 1996.,
  IEEE International Conference on}, vol.~3.\hskip 1em plus 0.5em minus
  0.4em\relax IEEE, 1996, pp. 1765--1770.

\bibitem{larsen2004case}
J.~C. Larsen and N.~J. Ferrier, ``A case study in vision based neural network
  training for control of a planar, large deflection, flexible robot
  manipulator,'' in \emph{Intelligent Robots and Systems, 2004.(IROS 2004).
  Proceedings. 2004 IEEE/RSJ International Conference on}, vol.~3.\hskip 1em
  plus 0.5em minus 0.4em\relax IEEE, 2004, pp. 2924--2929.

\bibitem{wang2005improving}
D.~Wang and Y.~Bai, ``Improving position accuracy of robot manipulators using
  neural networks,'' in \emph{Instrumentation and Measurement Technology
  Conference, 2005. IMTC 2005. Proceedings of the IEEE}, vol.~2.\hskip 1em plus
  0.5em minus 0.4em\relax IEEE, 2005, pp. 1524--1526.

\bibitem{comaniciu2002meanshift}
D.~Comaniciu and P.~Meer, ``Mean shift: a robust approach toward feature space
  analysis,'' \emph{Pattern Analysis and Machine Intelligence, IEEE
  Transactions on}, vol.~24, no.~5, pp. 603--619, May 2002.

\bibitem{huang2011extreme}
G.-B. Huang, D.~H. Wang, and Y.~Lan, ``Extreme learning machines: a survey,''
  \emph{International Journal of Machine Learning and Cybernetics}, vol.~2,
  no.~2, pp. 107--122, 2011.

\bibitem{huang2012extreme}
G.-B. Huang, H.~Zhou, X.~Ding, and R.~Zhang, ``Extreme learning machine for
  regression and multiclass classification,'' \emph{Systems, Man, and
  Cybernetics, Part B: Cybernetics, IEEE Transactions on}, vol.~42, no.~2, pp.
  513--529, 2012.

\bibitem{deng2009regularized}
W.~Deng, Q.~Zheng, and L.~Chen, ``Regularized extreme learning machine,'' in
  \emph{Computational Intelligence and Data Mining, 2009. CIDM'09. IEEE
  Symposium on}.\hskip 1em plus 0.5em minus 0.4em\relax IEEE, 2009, pp.
  389--395.

\bibitem{laviola2003double}
J.~J. LaViola, ``Double exponential smoothing: an alternative to kalman
  filter-based predictive tracking,'' in \emph{Proceedings of the workshop on
  Virtual environments 2003}.\hskip 1em plus 0.5em minus 0.4em\relax ACM, 2003,
  pp. 199--206.

\end{thebibliography}

\end{document}